\documentclass[10pt,twocolumn,letterpaper]{article}

\usepackage[pagenumbers]{wacv}      
\usepackage{graphicx}
\usepackage{amsmath}
\usepackage{amssymb}
\usepackage{booktabs}
\usepackage{multirow}
\usepackage{algorithm}
\usepackage{listings}
\usepackage{caption}

\usepackage[pagebackref,breaklinks,colorlinks]{hyperref}

\usepackage[capitalize]{cleveref}
\crefname{section}{Sec.}{Secs.}
\Crefname{section}{Section}{Sections}
\Crefname{table}{Table}{Tables}
\crefname{table}{Tab.}{Tabs.}

\begin{document}

\title{SC-MIL: Supervised Contrastive Multiple Instance Learning for Imbalanced Classification in Pathology}

\author{Dinkar Juyal\\
PathAI Inc\\
Boston, USA\\
{\tt\small dinkar.juyal@pathai.com}
\and
Siddhant Shingi$^{*}$\\
University of Massachusetts\\
Amherst, USA
\and
Syed Ashar Javed\\
PathAI Inc\\
\and
Harshith Padigela\\
PathAI Inc\\
\and
Chintan Shah\\
PathAI Inc\\
\and
Anand Sampat\\
PathAI Inc\\
\and
Archit Khosla\\
PathAI Inc\\
\and
John Abel\\
PathAI Inc\\
\and
Amaro Taylor-Weiner\\
PathAI Inc\\
}

\maketitle
\def\thefootnote{*}\footnotetext{Work done during internship at PathAI}
\begin{abstract}
Multiple Instance learning (MIL) models have been extensively used in pathology to predict biomarkers and risk-stratify patients from gigapixel-sized images. Machine learning problems in medical imaging often deal with rare diseases, making it important for these models to work in a label-imbalanced setting. In pathology images, there is another level of imbalance, where given a positively labeled Whole Slide Image (WSI), only a fraction of pixels within it contribute to the positive label. This compounds the severity of imbalance and makes imbalanced classification in pathology challenging. Furthermore, these imbalances can occur in out-of-distribution (OOD) datasets when the models are deployed in the real-world. We leverage the idea that decoupling feature and classifier learning can lead to improved decision boundaries for label imbalanced datasets. To this end, we investigate the integration of supervised contrastive learning with multiple instance learning (SC-MIL). Specifically, we propose a joint-training MIL framework in the presence of label imbalance that progressively transitions from learning bag-level representations to optimal classifier learning. We perform experiments with different imbalance settings for two well-studied problems in cancer pathology: subtyping of non-small cell lung cancer and subtyping of renal cell carcinoma. SC-MIL provides large and consistent improvements over other techniques on both in-distribution (ID) and OOD held-out sets across multiple imbalanced settings.
\end{abstract}

\section{Introduction}
\label{sec:intro}

Pathology is the microscopic study of tissue and a key component in medical diagnosis and drug development \cite{walk2009role}. The digitization of tissue slides, resulting in whole slide images (WSIs), has made pathology data more accessible for quantitative analysis. However, the large size (billions of pixels) and information density (hundreds of thousands of cells and heterogeneous tissue organization) of WSIs make manual analysis challenging \cite{niazi2019digital,javed2022rethinking}, highlighting the need for machine learning (ML) approaches \cite{wang2016deep,campanella2019clinical,bosch2021machine,SelfTrainingML4H,bulten2021artificial, chen2022scaling, MADABHUSHI2016170, bejnordi2017lymph}. ML techniques have been used for predicting a patient's clinical characteristics from a WSI. These models predict a label or score for the entire WSI, referred to as a slide-level prediction. Traditional approaches for handling large WSIs include the use of hand-engineered representations or breaking the slide into thousands of smaller patches \cite{diao2021human}. Both of these approaches require pixel or patch level annotations which are costly. To overcome the need for patch level labels, multiple instance learning (MIL) \cite{maron1997framework} has been applied to pathology by treating patches from slides as instances that form a bag, with a slide-level label associated with each bag. The MIL framework thus provides an end-to-end learning approach for problems in pathology. 

\begin{figure}
  \centering
    \includegraphics[scale=0.25]{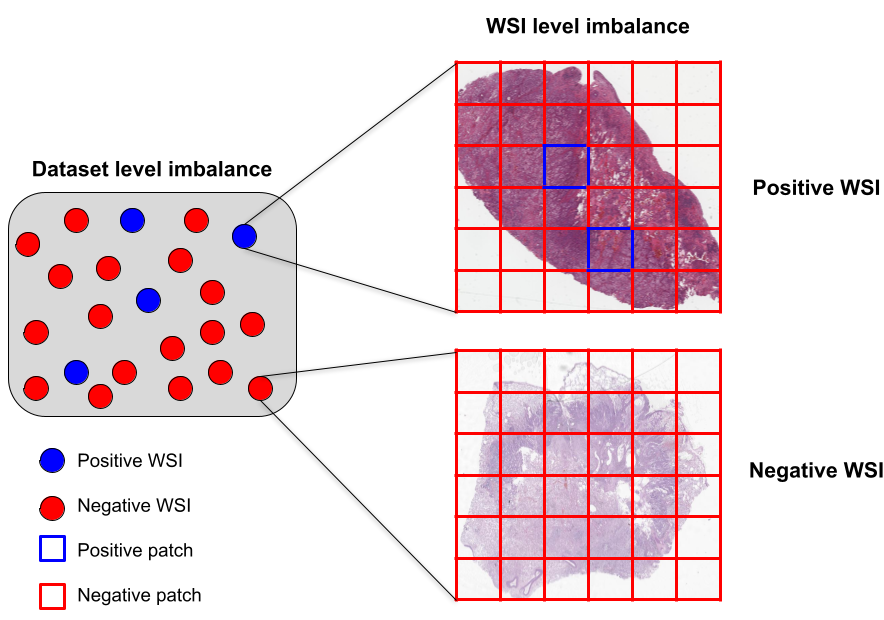}
    
    \caption{Label imbalance in histopathology domain is present at two levels - Dataset and Whole Slide Image (WSI). In datasets, imbalance arises from different prevalence rates of diseases. For a given WSI with positive label, only a small subset of patches contribute to that positive label. This compounds the severity of imbalance, making imbalanced classification in pathology challenging.}
    \label{sc-mil-problem-motivation}
\end{figure}

\begin{figure*}
  \centering
    \includegraphics[scale=0.45]{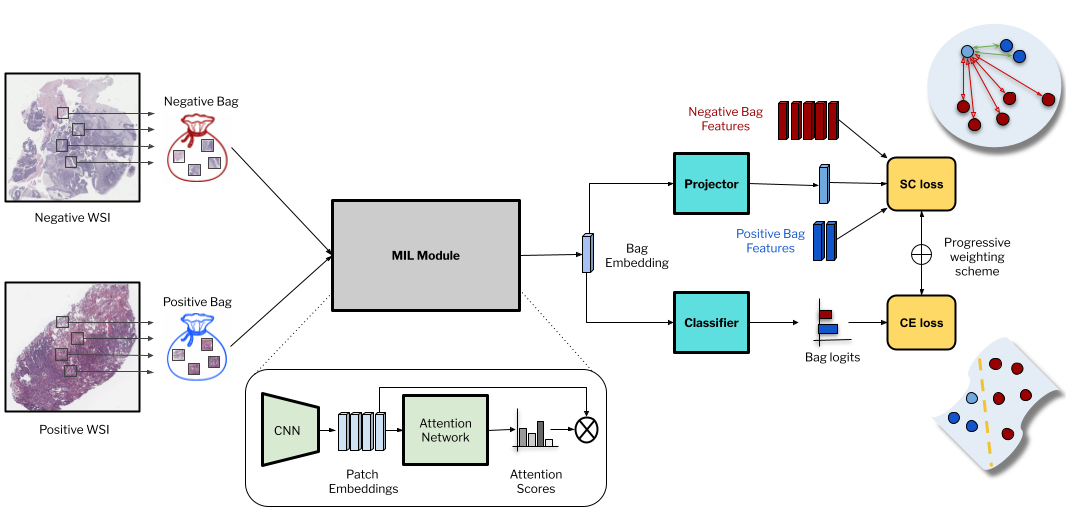}
    \rule{.7\linewidth}{0pt}
    
    \caption{SC-MIL integrates supervised contrastive learning into the MIL framework. The model performs joint feature and classifier learning on bag representations computed using an attention-based aggregation on patches. The training objective transitions progressively from a contrastive to a classification loss.}
    \label{sc-mil}
\end{figure*}

Label distribution in real-world settings can vary considerably depending on factors such as disease prevalence, population characteristics and the hospital or laboratory of origin. For example, a dataset of WSIs from a diagnostic lab may have a different class distribution compared to a dataset from a clinical trial enriched for certain disease characteristics. In fact, label imbalance in pathology datasets exists at both dataset and WSI level as shown in Figure \ref{sc-mil-problem-motivation}. MIL models should be robust to variations in label distribution to succeed in clinical applications and maintain physician trust.  Different approaches have been proposed to deal with label imbalance, ranging from data resampling (oversampling of minority classes or undersampling of majority classes) \cite{more2016surveyresample,mera2015bagoversample}, loss reweighting \cite{ren2018LearnReweigh}, selective enrichment of minority classes in image or feature space \cite{chou20120remix}, decoupling representation learning from classification \cite{zhou2020bbn}, and custom loss functions \cite{Cao2019ldam}.

Contrastive learning aims to learn representations that maximize the agreement among positive instances, e.g., different augmentations of the same image, and minimize the agreement with negative instances, e.g., other images in the dataset \cite{Ciga2020SelfSC}. In supervised contrastive learning (SCL) \cite{Khosla2020scl}, the contrastive loss formulation incorporates label information by treating all instances within the same class as positive examples for a given image. SCL adopts a two-stage learning technique where a feature extractor is learned in the first stage using a contrastive loss, followed by learning a classifier using the cross-entropy loss in the second stage.

This work proposes SC-MIL: a novel MIL technique to tackle label imbalance in pathology, that integrates SCL into the MIL framework. We take inspiration from prior work \cite{kang2021balancedfeat,Graf2021DissectingSC} which shows that a) contrastive loss learns balanced feature spaces (i.e., feature spaces with similar inter-class separation for all classes) compared to cross-entropy, and b) this balance is positively related to performance across imbalanced settings. Additionally, we use a smooth transition from feature learning to classifier learning in the course of training, which allows the model to learn a more discriminative latent space, aiding in imbalanced classification \cite{Wang2021HybridContrastive}. In the MIL setting, labels are only available for a bag (i.e., a collection of patches) and not individual patches. Applying SCL to patch features assumes assigning a bag-label to individual patches. However, a single patch might not have any information about the WSI label. For example, a malignant WSI might have many patches which contain only normal tissue. 
This motivates our bag-level formulation of SC-MIL where contrastive loss is applied to the bag features. Feature learning with bag-level contrastive loss tackles dataset imbalance, while the multiple instance formulation addresses imbalance within a WSI.  The contributions of this work are as follows:
\begin{enumerate}
    \item We tackle the problem of label imbalance by proposing a formulation that extends SCL to the MIL setting. We investigate two training strategies for optimal feature and classifier learning with SC-MIL.
    \item We conduct an extensive study on the performance of this technique across different degrees of label imbalance on two open-source datasets: subtyping in non-small cell lung cancer (NSCLC) and renal cell carcinoma (RCC). We compare this to previous state-of-the-art methods used for label imbalance and demonstrate the effectiveness of using SC-MIL over these methods.
    \item We show substantial performance improvements with SC-MIL on OOD data across multiple degrees of label imbalance, making a strong case for its utility in real-world deployment scenarios.

\end{enumerate}

\begin{table}
    \centering
        \caption{Training data-distribution of TCGA RCC sub-typing across imbalance ratios}
        \begin{tabular*}{\linewidth}{@{\extracolsep{\stretch{1}}}@{}cccc@{}}
            \toprule
            \multicolumn{1}{c}{\multirow{2}{*}{Classes}} & \multicolumn{3}{c}{Imbalance Ratio} \\ \cmidrule(lr){2-4} 
            \multicolumn{1}{c}{}                         & 1          & 5          & 10        \\ \midrule
            KIRC                                         & 96         & 205        & 240       \\
            KIRP                                         & 96         & 41         & 24        \\
            KICH                                         & 96         & 41         & 24        \\ \midrule
            Total                                        & 288        & 287        & 288      \\ \bottomrule
            \end{tabular*}%
           
        \label{tab:rcc_data}

\end{table}

\begin{table}

    \centering
    \caption{Training data-distribution of TCGA NSCLC sub-typing across imbalance ratios}
    
    \begin{tabular*}{\linewidth}{@{\extracolsep{\stretch{1}}}@{}cccc@{}}
        \toprule
        \multicolumn{1}{c}{\multirow{2}{*}{Classes}} & \multicolumn{3}{c}{Imbalance Ratio} \\ \cmidrule(lr){2-4} 
        \multicolumn{1}{c}{}                         & 1          & 5          & 10        \\ \midrule
        LUAD                                        & 158        & 265        & 290       \\
        LUSC                                     & 158        & 53         & 29        \\  \midrule
        Total                                        & 316        & 318        & 319      \\ \bottomrule
        \end{tabular*}%
        
        \label{tab:nsclc_data}
    
\end{table}

\section{Supervised Contrastive Multiple Instance Learning}

\subsection{Multiple Instance Learning}
MIL is a weakly supervised learning approach that allows learning and making predictions on a group of instances. Unlike supervised learning, the MIL framework only requires labels for the group of instances, called a bag, but not the individual instances. This is valuable in the context of pathology, where a collection of patches from a WSI can be treated as a bag and this allows learning of slide-level predictors without the need for fine-grained patch-level annotations. For pooling of patches, a learnt attention based aggregation scheme 
\cite{ilse2018attention} has been shown to be effective and is commonly used in end-to-end pathology models.

In the binary case, a bag is considered positive if it has at least one positive instance and negative if there are none. Given a set of instances $X = \{x_0, x_1,\dots\, x_n\}$, the MIL prediction $p(X)$ is 
\begin{equation} \label{eq:1}
p(X) = a(f(x_0), f(x_1), ....., f(x_n))
\end{equation}
where $f$ is an encoder for instances, $a$ is a permutation-invariant aggregator, mapping from feature space to the prediction space. 
Learnt aggregation functions like AttentionMIL and its variants DSMIL \cite{li2021dual}, CLAM \cite{lu2021data}, TransMIL \cite{shao2021transmil}, AdditiveMIL \cite{javed2022additive} have shown significant improvements over heuristic aggregators like Max or Mean in various tasks \cite{ilse2018attention}. We will focus on the AttentionMIL (also referred to as ABMIL) formulation for our discussion.

The aggregator function $a$ in AttentionMIL has two components. An attention module $m$ induces a soft-attention $\alpha_i$ over the instances and computes an attention weighted aggregation of instance features to generate the bag embedding $b(X)$. A classifier $h$ maps the bag feature to the bag prediction.
\begin{equation} \label{eq:3}
p(X) = h(b(X))
\end{equation}
\begin{equation}
b(X) = m(f(x_0), f(x_1), ....., f(x_n)) = \sum_{i=0}^{i=n} \alpha_i f(x_i)
\end{equation}
\begin{equation}
\alpha_i = softmax(\phi_m(x_i))
\end{equation}
where $\phi_m$ is a neural network with a non-linear activation.

\begin{algorithm}[t]
\caption{SC-MIL Pseudocode, PyTorch-like}
\label{alg:code}
\definecolor{codeblue}{rgb}{0.25,0.5,0.5}
\definecolor{codekw}{rgb}{0.85, 0.18, 0.50}
\lstset{
  backgroundcolor=\color{white},
  basicstyle=\fontsize{8.0pt}{8.0pt}\ttfamily\selectfont,
  columns=fullflexible,
  breaklines=true,
  captionpos=b,
  commentstyle=\fontsize{8.0pt}{8.0pt}\color{codeblue},
  keywordstyle=\fontsize{8.0pt}{8.0pt}\color{codekw},
}
\begin{lstlisting}[language=python]
# f: patch level feature extractor
# m: attention module
# g: projector MLP
# h: classifier
# X: bag of patches
# Y: bag label

# load a bag X=[x_1, ..., x_n] with n patches
for X in loader:
    # patch level embeddings, n-by-d
    E = f(X) 
    # attention weights, n-by-1
    attn_wts = m(E)
    # bag level embedding, d-by-1
    B = Sum(attn_wts*E) 

    # projected bag embedding
    Z = norm(g(B)) 
    # bag level predictions
    P = h(B) 

    # t_i: the current iteration
    # t: total number of iterations
    beta_t = 1 - t_i/t
    L_scl = scl_loss(Z, other Z_i in minibatch)
    L_ce = cross_entropy_loss(P, Y)
    Loss = beta_t*L_scl + (1-beta_t)*L_ce

    Loss.backward()

\end{lstlisting}
\end{algorithm}

\begin{center}
\begin{table*}[]
    \caption{Comparison of SC-MIL with other label imbalance techniques on TCGA-RCC test set for RCC subtyping (RS - Random Sampling, CB - Class Balanced)}
    \resizebox{\textwidth}{!}{%
    \begin{tabular}{ccccccc@{\hskip 0.08in}cccccc}
    \toprule
    Dataset         & \multicolumn{6}{c}{TCGA-RCC}  \\ \cmidrule(lr){1-1} \cmidrule(lr){2-7}
    Imbalance Ratio & \multicolumn{2}{c}{1}                                                      & \multicolumn{2}{c}{5}                                                      & \multicolumn{2}{c}{10}  \\ \cmidrule(lr){1-1} \cmidrule(lr){2-3} \cmidrule(lr){4-5} \cmidrule(lr){6-7}
        
        Metric (\%)          & \multicolumn{1}{c}{F1}              & \multicolumn{1}{c}{AUC}           & \multicolumn{1}{c}{F1}              & \multicolumn{1}{c}{AUC}           & \multicolumn{1}{c}{F1}              & AUC       \\ \midrule
        ABMIL-RS          & \multicolumn{1}{c}{87.17$\pm$2.03}          & \multicolumn{1}{c}{96.13$\pm$0.96}          & \multicolumn{1}{c}{83.40$\pm$2.52}            & \multicolumn{1}{c}{95.6$\pm$0.99}          & \multicolumn{1}{c}{78.23$\pm$2.82}          & 93.26$\pm$1.39        \\
        ABMIL-CB          & \multicolumn{1}{c}{89.49$\pm$2.12}          & \multicolumn{1}{c}{97.42$\pm$0.85}          & \multicolumn{1}{c}{84.61$\pm$2.84}          & \multicolumn{1}{c}{93.93$\pm$1.60}          & \multicolumn{1}{c}{73.63$\pm$2.83}          & 95.05$\pm$0.84       \\ \midrule
        LDAM-DRW        & \multicolumn{1}{c}{89.35$\pm$1.92}          & \multicolumn{1}{c}{97.65$\pm$0.66}          & \multicolumn{1}{c}{83.50$\pm$2.45}          & \multicolumn{1}{c}{94.96$\pm$1.23}          & \multicolumn{1}{c}{80.66$\pm$2.93}          & 93.08$\pm$1.54    \\ \midrule
        SC-MIL-RS          & \multicolumn{1}{c}{88.67$\pm$2.21}          & \multicolumn{1}{c}{\textbf{98.13$\pm$0.67}} & \multicolumn{1}{c}{\textbf{86.30$\pm$2.03}} & \multicolumn{1}{c}{\textbf{96.83$\pm$0.66}} & \multicolumn{1}{c}{\textbf{87.42$\pm$2.07}} & \textbf{96.40$\pm$0.94}     \\ 
        SC-MIL-CB          & \multicolumn{1}{c}{\textbf{90.13$\pm$2.17}} & \multicolumn{1}{c}{97.98$\pm$0.65}          & \multicolumn{1}{c}{85.53$\pm$2.55}          & \multicolumn{1}{c}{96.69$\pm$0.82}          & \multicolumn{1}{c}{81.34$\pm$2.66}          & 96.35$\pm$1.11   \\ \bottomrule
    \end{tabular}%
    }

\label{tab:rcc_results}
\end{table*}
\end{center}

\begin{center}
\begin{table*}[]
    \caption{Comparison of SC-MIL with other label imbalance techniques on OOD-RCC test set for RCC subtyping (RS - Random Sampling, CB - Class Balanced)}
    \resizebox{\textwidth}{!}{%
    \begin{tabular}{ccccccc}
    \toprule
    Dataset         & \multicolumn{6}{c}{OOD-RCC}  \\ \cmidrule(lr){1-1} \cmidrule(lr){2-7} 
    Imbalance Ratio & \multicolumn{2}{c}{1}                                                      & \multicolumn{2}{c}{5}                                                      & \multicolumn{2}{c}{10}                                \\ \cmidrule(lr){1-1} \cmidrule(lr){2-3} \cmidrule(lr){4-5} \cmidrule(lr){6-7}
    Metric (\%)          & \multicolumn{1}{c}{F1}              & \multicolumn{1}{c}{AUC}           & \multicolumn{1}{c}{F1}              & \multicolumn{1}{c}{AUC}           & \multicolumn{1}{c}{F1}              & AUC          \\ \midrule
        ABMIL-RS         & \multicolumn{1}{c}{74.20$\pm$1.91}          & \multicolumn{1}{c}{93.88$\pm$1.09}          & \multicolumn{1}{c}{73.69$\pm$2.55}          & \multicolumn{1}{c}{91.15$\pm$1.56}           & \multicolumn{1}{c}{70.23$\pm$3.42}          & 87.74$\pm$2.10          \\
        ABMIL-CB      & \multicolumn{1}{c}{77.33$\pm$2.55}          & \multicolumn{1}{c}{92.79$\pm$1.48}          & \multicolumn{1}{c}{72.31$\pm$2.33}          & \multicolumn{1}{c}{89.49$\pm$1.65}          & \multicolumn{1}{c}{71.38$\pm$2.88}          & 91.82$\pm$1.47          \\ \midrule
        LDAM-DRW   & \multicolumn{1}{c}{78.97$\pm$2.45}          & \multicolumn{1}{c}{93.69$\pm$1.38}          & \multicolumn{1}{c}{73.47$\pm$2.52}          & \multicolumn{1}{c}{88.62$\pm$1.97}          & \multicolumn{1}{c}{72.42$\pm$2.51}          & 91.94$\pm$1.52          \\ \midrule
        SC-MIL-RS  & \multicolumn{1}{c}{\textbf{81.94$\pm$2.39}} & \multicolumn{1}{c}{\textbf{94.84$\pm$1.24}} & \multicolumn{1}{c}{\textbf{81.87$\pm$2.54}} & \multicolumn{1}{c}{\textbf{93.42$\pm$1.43}} & \multicolumn{1}{c}{\textbf{80.91$\pm$2.24}} & 92.57$\pm$1.41         \\ 
        SC-MIL-CB   & \multicolumn{1}{c}{76.81$\pm$2.31}          & \multicolumn{1}{c}{93.78$\pm$7.83}          & \multicolumn{1}{c}{79.04$\pm$2.34}          & \multicolumn{1}{c}{92.56$\pm$1.51}          & \multicolumn{1}{c}{79.04$\pm$2.34}          & \textbf{92.56$\pm$1.51} \\ \bottomrule
    \end{tabular}%
    }

\label{tab:rcc_results_ood}
\end{table*}
\end{center}

\begin{center}
\begin{table*}[]
\caption{Comparison of SC-MIL with other label imbalance techniques on TCGA-NSCLC test set for NSCLC subtyping (RS - Random Sampling, CB - Class Balanced)}
    \resizebox{\textwidth}{!}{%
    \begin{tabular}{ccccccc}
    \toprule
    Dataset         & \multicolumn{6}{c}{TCGA-NSCLC} \\ \cmidrule(lr){1-1} \cmidrule(lr){2-7}
    Imbalance Ratio & \multicolumn{2}{c}{1}                                                      & \multicolumn{2}{c}{5}                                                      & \multicolumn{2}{c}{10}                               \\ \cmidrule(lr){1-1} \cmidrule(lr){2-3} \cmidrule(lr){4-5} \cmidrule(lr){6-7} 
    Metric (\%)         & \multicolumn{1}{c}{F1}              & \multicolumn{1}{c}{AUC}           & \multicolumn{1}{c}{F1}              & \multicolumn{1}{c}{AUC}           & \multicolumn{1}{c}{F1}              & AUC           \\ \midrule
    ABMIL-RS          & \multicolumn{1}{c}{82.07$\pm$2.5}          & \multicolumn{1}{c}{91.36$\pm$1.76}          & \multicolumn{1}{c}{81.62$\pm$2.62}            & \multicolumn{1}{c}{89.84$\pm$2.03}          & \multicolumn{1}{c}{77.35$\pm$3.18}          & 89.68$\pm$2.14           \\ 
    ABMIL-CB          & \multicolumn{1}{c}{83.23$\pm$2.45}          & \multicolumn{1}{c}{91.81$\pm$1.76}          & \multicolumn{1}{c}{82.61$\pm$2.72}          & \multicolumn{1}{c}{90.13$\pm$1.99}          & \multicolumn{1}{c}{78.0$\pm$2.95}          & 88.57$\pm$2.18        \\ \midrule
    LDAM-DRW        & \multicolumn{1}{c}{85.8$\pm$2.28}          & \multicolumn{1}{c}{91.9$\pm$1.78}          & \multicolumn{1}{c}{80.91$\pm$2.91}          & \multicolumn{1}{c}{89.89$\pm$2.13}          & \multicolumn{1}{c}{81.56$\pm$2.79}          & 89.14$\pm$2.15      \\ \midrule
    SC-MIL-RS          & \multicolumn{1}{c}{\textbf{87.65$\pm$2.21}}  & \multicolumn{1}{c}{\textbf{94.81$\pm$1.34}}  & \multicolumn{1}{c}{86.66$\pm$2.34}        & \multicolumn{1}{c}{\textbf{92.45$\pm$1.78}} & \multicolumn{1}{c}{\textbf{84.05$\pm$2.65}} & \textbf{91.64$\pm$1.81}  \\ 
    SC-MIL-CB          & \multicolumn{1}{c}{84.85$\pm$2.52}         & \multicolumn{1}{c}{93.14$\pm$1.65}          & \multicolumn{1}{c}{\textbf{87.02$\pm$2.29}}   & \multicolumn{1}{c}{92.21$\pm$1.87}          & \multicolumn{1}{c}{80.37$\pm$3.02}          & 90.96$\pm$1.93      \\ \bottomrule
    \end{tabular}%
    }
    \label{tab:nsclc_results}
\end{table*}
\end{center}

\begin{center}
\begin{table*}[]
    \caption{Comparison of SC-MIL with other label imbalance techniques on OOD-NSCLC test set for NSCLC subtyping (RS - Random Sampling, CB - Class Balanced)}
        \resizebox{\textwidth}{!}{%
        \begin{tabular}{ccccccc}
        \toprule
        Dataset         & \multicolumn{6}{c}{OOD-NSCLC}  \\  \cmidrule(lr){1-1} \cmidrule(lr){2-7}
        Imbalance Ratio & \multicolumn{2}{c}{1}   & \multicolumn{2}{c}{5}    & \multicolumn{2}{c}{10}  \\  \cmidrule(lr){1-1} \cmidrule(lr){2-3} \cmidrule(lr){4-5} \cmidrule(lr){6-7} 
        Metric (\%)         & \multicolumn{1}{c}{F1}              & \multicolumn{1}{c}{AUC}           & \multicolumn{1}{c}{F1}              & \multicolumn{1}{c}{AUC}           & \multicolumn{1}{c}{F1}              & AUC           \\ \midrule
        ABMIL-RS    & \multicolumn{1}{c}{71.03$\pm$5.34}          & \multicolumn{1}{c}{92.42$\pm$2.33}          & \multicolumn{1}{c}{19.82$\pm$7.91}          & \multicolumn{1}{c}{75.95$\pm$4.24}           & \multicolumn{1}{c}{12.34$\pm$6.5}         & 88.61$\pm$2.64          \\ 
        ABMIL-CB    & \multicolumn{1}{c}{36.3$\pm$7.97}          & \multicolumn{1}{c}{92.54$\pm$1.93}          & \multicolumn{1}{c}{15.58$\pm$7.04}          & \multicolumn{1}{c}{77.25$\pm$4.41}          & \multicolumn{1}{c}{8.14$\pm$5.57}         & 76.71$\pm$4.01          \\ \midrule
        LDAM-DRW   & \multicolumn{1}{c}{61.63$\pm$6.81}          & \multicolumn{1}{c}{90.31$\pm$2.5}          & \multicolumn{1}{c}{26.28$\pm$8.11}          & \multicolumn{1}{c}{89.57$\pm$2.56}          & \multicolumn{1}{c}{28.85$\pm$8.02}          & 88.36$\pm$2.89         \\ \midrule
        SC-MIL-RS  & \multicolumn{1}{c}{\textbf{76.46$\pm$5.27}} & \multicolumn{1}{c}{\textbf{93.64$\pm$2.45}} &       \multicolumn{1}{c}{37.68$\pm$7.75} & \multicolumn{1}{c}{91.58$\pm$2.02}                   & \multicolumn{1}{c}{\textbf{41.5$\pm$8.23}} & \textbf{92.97$\pm$1.95}          \\ 
        SC-MIL-CB   & \multicolumn{1}{c}{58.82$\pm$6.68}          & \multicolumn{1}{c}{84.83$\pm$3.08}          & \multicolumn{1}{c}{\textbf{49.65$\pm$8.26}}   & \multicolumn{1}{c}{\textbf{94.23$\pm$1.95}}          & \multicolumn{1}{c}{29.06$\pm$8.2}          & 79.04$\pm$4.4 \\ \bottomrule
        \end{tabular}%
    }
    \label{tab:nsclc_results_ood}
\end{table*}
\end{center}

\subsection{SC-MIL: Supervised Contrastive Multiple Instance Learning} \label{sec:sc_mil}
SCL \cite{Khosla2020scl} proposes a way to leverage contrastive learning and incorporate supervision. It learns instance representations by pulling instances from same class together and those from different classes apart in the representation space. In MIL, we can use SCL for learning either instance or bag representations. Considering we only have labels for bags and not individual instances, using SCL to learn instance representations needs using bag labels as instance labels, thus introducing label noise and breaking the MIL assumption. Instead, we use SCL to learn bag representations.

Specifically, given a set of instances for a bag $X_i = \{x_0, x_1,\dots\, x_n\}$, we compute the bag representation $b(X_i)$ using the MIL formulation, where $i$ denotes the index of a bag in a given batch. We now use a non-linear multi-layer perceptron $g$ to generate the projection $z_i$ for the bag representation. We then compute the SCL loss for MIL $\mathcal{L}_{SCL}$ as follows:
\begin{equation}
    z_i = g(b(X_i)) 
\end{equation}
\begin{equation}
    \mathcal{L}_{SCL}=\sum_i -\frac{1}{|\textbf{P}_i^{+}|}\sum_{z_j \in \textbf{P}_i^{+}}\log\frac{\exp(z_i \cdot z_j/\tau)}{\sum_{z_k \in \textbf{B}_i}\exp(z_i \cdot z_k/\tau)}
\end{equation}
where $\textbf{P}_i^{+}$ denotes the positive bags sharing the same class label as bag $z_i$ and $\textbf{B}_i$ is the set of all bags in the batch excluding bag $z_i$.

 Curriculum-based feature and classifier learning using both contrastive and cross entropy losses has been shown to be effective in long-tailed image classification \cite{Kang2020Decoupling}. We apply the same approach to the MIL setting at a bag level. For classifier learning, we use the cross-entropy loss. The classifier branch projects the bag embedding $b(X)$ to the prediction $p(X)$ as shown in Equation \ref{eq:3} and uses cross entropy $\mathcal{L}_{CE}$ to learn the classifier:
\begin{equation}
    \mathcal{L}_{SC-MIL} = \beta_t \mathcal{L}_{SCL} + (1 - \beta_t) \mathcal{L}_{CE}
\end{equation}
 where the weight $\beta_t \in [0,1]$ is decayed through the course of training iterations $t$ using a curriculum to gradually transition from feature to classifier learning.

\begin{figure}
  \centering
    \includegraphics[scale=0.2]{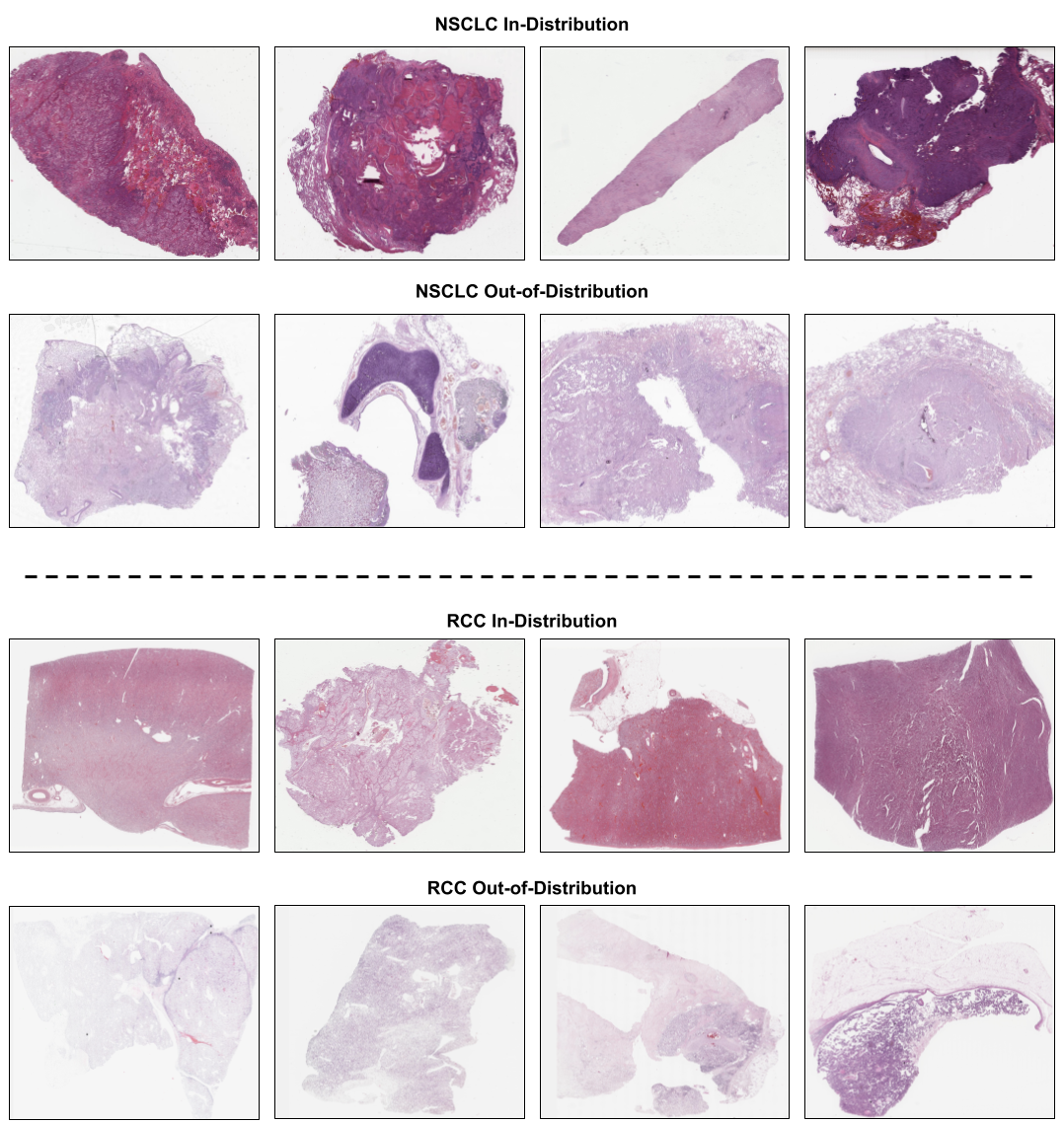}
    
    \caption{ Visual comparison of in-distribution (ID) and out-of-distribution (OOD) WSIs from the cancer subtyping datasets. The first and third rows show in-distribution TCGA WSIs from NSCLC and RCC respectively. The second and fourth rows show WSIs procured from a different lab site and scanner. We can see the variations in tissue preparation and scanning which lead to significant drops in performance.}

    \label{fig:id-and-ood}
\end{figure}

\section{Experiments and Results} 
We first introduce the datasets used for experimentation. We describe the mechanism of simulating different degrees of imbalance in these datasets while ensuring that the total number of samples remains consistent. We then discuss results on all datasets using SC-MIL and other baselines. Finally, we present ablation studies to understand the tradeoffs made in terms of training supervised contrastive loss with cross-entropy jointly vs sequentially, and the impact of hyperparameters. 

\begin{figure*}
  \centering
    \includegraphics[scale=0.68]{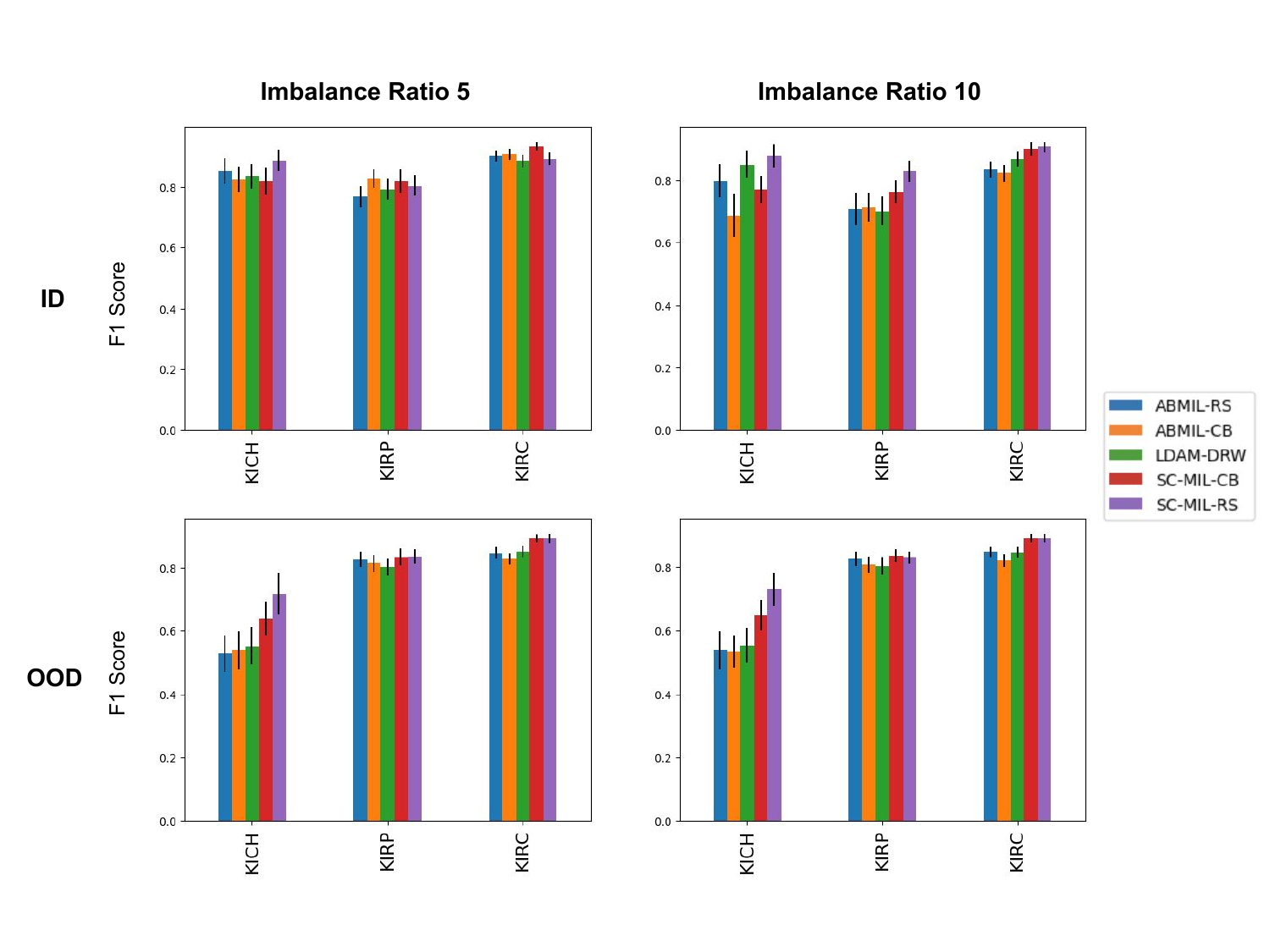}
    \caption{Class-wise F1 score comparison for RCC subtyping: SC-MIL outperforms other methods across different imbalance ratios. The performance gains are higher on minority classes, and they increase on moving from ID to OOD datasets.}
    \label{fig:classwise-f1-scores-rcc}
\end{figure*}

\subsection{Datasets and Setup} 
We considered two datasets from The Cancer Genome Atlas (TCGA) \cite{weinstein2013cancer}  - prediction of cancer subtypes in non-small cell lung carcinoma (NSCLC) and renal cell carcinoma (RCC). TCGA-NSCLC contains a total of 1002 WSIs stained with H\&E, 538 of which were collected from patients with the adenocarcinoma histologic subtype (LUAD) and 464 from squamous cell carcinoma (LUSC). TCGA-RCC contains 948 WSIs with three histologic subtypes: 158 WSIs with the label chromophobe RCC (KICH), 504 WSIs belonging to clear cell RCC (KIRC), and 286 to papillary RCC (KIRP).

We performed a label-stratified split of both datasets while ensuring there is no leakage of case information (i.e., combination of tissue source site and study participant) across splits. The splitting ratio was 60:15:25 (train:val:test); other clinical or sample characteristics were not used during splitting. To simulate varying degrees of label imbalance, we sampled WSIs from the available classes to generate imbalance in the train set, while the heldout sets were kept the same. In line with previous works \cite{Cao2019ldam,Wang2021HybridContrastive}, we used imbalance ratio $\rho = \frac{\max_i\{n_i\}}{\min_i\{n_i\}}$ which denotes the ratio of number of examples of the majority class to the minority class. We experimented with imbalance ratios of 1, 5 and 10. We ensured that the number of training examples remained consistent across different imbalance ratios to remove any confounding effect of the number of data points and to enable comparison of model performance across imbalance ratios. Since there were three classes in TCGA-RCC, the two classes with least number of samples (KIRP and KICH) were treated as minority classes. The details of the resulting dataset composition is shared in Table \ref{tab:rcc_data} and \ref{tab:nsclc_data}.

We also deployed all models on two OOD datasets collected from different patient populations and having different sample characteristics for NSCLC and RCC. These OOD datasets are acquired from other laboratories using varying image acquisition and processing steps resulting in visual differences from their TCGA counterparts. OOD NSCLC has 162 LUAD and 45 LUSC WSIs, while OOD RCC has 254 KIRC, 134 KIRP and 46 KICH WSIs. Example images comparing ID and OOD datasets are shared in Figure \ref{fig:id-and-ood}.

\subsection{Implementation Details} 
We trained five models: a baseline AttentionMIL model with random sampling (ABMIL-RS) and class balanced sampling (ABMIL-CB), a version using label-distribution-aware margin loss with deferred reweighting (LDAM-DRW \cite{Cao2019ldam}, previously shown to be successful for addressing label imbalance in single instance classification), and our proposed SC-MIL with random (SC-MIL-RS) and class balanced sampling (SC-MIL-CB). Non-overlapping patches of size $224 \times 224$ pixels were selected from tissue regions (using a separate model which masks background and artifacts) at a resolution of 1 micron per pixel. We extracted 1.45 million patches from TCGA-NSCLC and 768k patches from TCGA-RCC. Bag sizes (number of patches in a bag) varied from 24 to 1500 patches and batch sizes (number of bags in a batch) varied from 8 to 32. Augmentations applied included color-based augmentations (random grayscaling, HSV transforms), gaussian blur and sharpening, horizontal and vertical flips, center crops. Augmentation related parameters were kept consistent across all techniques. An ImageNet-pretrained ShuffleNet \cite{shufflenet} was used to extract features from input patches. All models were trained end-to-end with the Adam optimizer and a learning rate of $1$e-$4$. SC-MIL models were trained with a temperature $\tau = 1$, and the training was performed jointly with cross entropy with a linear curriculum as described in Section \ref{sec:sc_mil}, with $\beta_t = 1$ at the start of training. For inference, patches were exhaustively sampled from a WSI and the majority prediction across bags was selected as the WSI-level prediction. For RCC, macro-averaged F1 score and macro-average of 1-vs-rest AUROC was computed.  Training and inference was performed on Quadro RTX 8000 GPUs using PyTorch v1.11 and CUDA 10.2. The training time for SC-MIL was comparable with other techniques (10-14 GPU hours). 

\begin{table}
\centering
\caption{Comparison of Patch vs Bag-level SCL on TCGA-RCC subtyping. All comparisons used SC-MIL-RS training and $\tau=1$}

\begin{tabular*}{\linewidth}{@{\extracolsep{\stretch{1}}}@{}cccc@{}}
    \toprule
    Ratio               & Method & F1    & AUC   \\ \midrule
    \multirow{2}{*}{1}  & SCL-RS     & \textbf{88.67$\pm$2.21} & \textbf{98.13$\pm$0.67} \\
                        & Patch SCL     & 82.87$\pm$2.81 & 97.33$\pm$0.82 \\ \midrule
    \multirow{2}{*}{5}  & SCL-RS     & \textbf{86.30$\pm$2.03} & \textbf{96.83$\pm$0.66} \\
                        & Patch SCL     & 79.68$\pm$2.87 & 94.71$\pm$1.05 \\ \midrule
    \multirow{2}{*}{10} & SCL-RS     & \textbf{87.42$\pm$2.07} & \textbf{96.40$\pm$0.94} \\
                        & Patch SCL     & 79.80$\pm$2.57 & 94.90$\pm$1.09 \\ \bottomrule
    \end{tabular*}%
\label{tab:patch_based_SCL}

\end{table}

\subsection{Experimental Results and Ablation Studies} 
\subsubsection{Comparison of SC-MIL with other techniques}
We compared the predictive performance of SC-MIL with other techniques across different imbalance ratios. Table \ref{tab:rcc_results} and \ref{tab:nsclc_results} show results on the NSCLC and RCC test sets respectively. SC-MIL outperforms other techniques across all imbalance ratios, and the difference is more pronounced at higher imbalance ratios. 
To further stress test these methods, we also deployed these models on independent OOD test datasets described above and the results are shown in Table \ref{tab:rcc_results_ood} and \ref{tab:nsclc_results_ood}. We found that baseline model performance dropped notably across imbalance ratios, highlighting the difficulty in generalization, and the tendency of these models to overfit in an imbalanced setting. Performance improvements using SC-MIL persist in this OOD setting. In Figure \ref{fig:classwise-f1-scores-rcc} we show the performance of different techniques across all classes in RCC in both ID and OOD setting, demonstrating the relative performance gain in each class.

\subsubsection{Patch vs bag based SC-MIL}
We conduct an experiment with a modification of SC-MIL architecture, where the supervised contrastive loss is applied on patch level embeddings instead of bag level embeddings. In Section \ref{sec:sc_mil}, we theorized that naively assigning the bag level label to instances and then applying supervised contrastive loss will result in incorrect label assignment. We show the results of training with such a scheme in Table \ref{tab:patch_based_SCL}. Patch level SC-MIL has inferior performance and higher variance as compared to our formulation. We also observe that the performance gap between the two models increases with increasing imbalance ratio, providing evidence that our bag-level formulation is more robust to the compounding effect of label imbalance in pathology.

\subsubsection{Impact of sampling} We found that SC-MIL with random sampling performs better than class balanced sampling in most cases. We hypothesize that this is due to reduced diversity in the feature space as a side effect of oversampling the minority classes or under sampling the majority class when using class-balanced sampling, which ultimately hurts performance by interfering with feature learning \cite{Wang2021HybridContrastive}. 

\subsubsection{Impact of temperature}
We experimented with temperature values of $\tau \in \{0.1, 0.5, 1\}$ and found that the models are generally robust to temperature changes as shown in Table \ref{tab:abalations}. We reason about this through two desirable properties of representations learned through contrastive learning: \textit{uniformity} in the hypersphere, i.e, inter-class separation and \textit{tolerance} to potential positives, i.e., intra-class similarity \cite{Wang2021understandingCL}. The former is favored by low values of temperature while higher values favor the latter. As shown in \cite{Wang2021understandingCL},  in problems with a larger number of classes, uniformity is harder to achieve and higher values of temperature harm feature quality. In contrast, we see that for RCC and NSCLC subtyping with 3 and 2 classes respectively, model performance is less sensitive to changes in temperature. 
\begin{table}
\centering
    \caption{Impact of temperature ($\tau$) on single-stage training SC-MIL-RS on TCGA-RCC subtyping.}
\begin{tabular*}{\linewidth}{@{\extracolsep{\stretch{1}}}@{}cccc@{}}
    \toprule
    Imb. Ratio          & Temp & F1     & AUC    \\ \midrule
    {1}  & 0.1  & 86.24$\pm$2.35 & 97.85$\pm$0.60 \\
                        & 0.5  & 87.88$\pm$2.37 & 97.87$\pm$0.72 \\
                        & 1.0  & \textbf{88.67$\pm$2.21}  & \textbf{98.13$\pm$0.67}  \\ \midrule
    {5}  & 0.1  &  86.14$\pm$2.61 & 97.04$\pm$0.96 \\
                        & 0.5  & \textbf{88.35$\pm$2.34}  & \textbf{97.49$\pm$0.68} \\
                    
                        & 1.0  & 86.30$\pm$2.03 & 96.83$\pm$0.66 \\ \midrule
    {10} & 0.1  & 85.60$\pm$2.37 & 96.09$\pm$0.97 \\
                        & 0.5  &84.85$\pm$2.30 & 96.06$\pm$0.78 \\
                        & 1.0  & \textbf{87.42$\pm$2.07} & \textbf{96.40$\pm$0.94}  \\ \bottomrule
    \end{tabular*}%

     \label{tab:abalations}

\end{table}

\begin{table}
\centering
    \caption{Comparison of one-stage vs two-stage training on TCGA-RCC subtyping. All comparisons used SC-MIL-RS training and $\tau=1$}

    \begin{tabular*}{\linewidth}{@{\extracolsep{\stretch{1}}}@{}cccc@{}}
    \toprule
    Ratio               & Stage & F1    & AUC   \\ \midrule
    {1}  & 1     & \textbf{88.67$\pm$2.21} & \textbf{98.13$\pm$0.67} \\
                        & 2     & 87.54$\pm$2.10 & 97.66$\pm$0.60 \\ \midrule
    {5}  & 1     & \textbf{86.30$\pm$2.03} & 96.83$\pm$0.66 \\
                        & 2     & 86.06$\pm$2.47 & \textbf{97.34$\pm$0.77} \\ \midrule
    {10} & 1     & \textbf{87.42$\pm$2.07} & \textbf{96.40$\pm$0.94} \\
                        & 2     & 85.26$\pm$2.56 & 96.01$\pm$0.91 \\ \bottomrule
    \end{tabular*}%
    
\label{tab:training_stages}

\end{table}

\subsubsection{Two-stage vs single-stage training}
We conducted an ablation by training models in a two-stage manner, with SCL loss in the first stage for feature learning followed by cross-entropy (CE) loss in the second stage. We see that single-stage SC-MIL model (joint SCL and CE training) performs better overall as shown in Table \ref{tab:training_stages}. This could be due to incompatible feature learning between SCL and CE stages in two-stage training. Using a smooth curriculum allows a gradual transition from feature learning to classifier learning, leading to superior performance.

\section{Conclusion} 
Label imbalance in pathology is a challenging problem owing to the highly skewed distribution of classes both at dataset and WSI level. We propose SC-MIL, a novel integration of supervised contrastive learning into the MIL framework to tackle this label imbalance problem. Experiments show our bag-level formulation to be superior to patch-level SC-MIL and other baselines across multiple degrees of label imbalance. Moreover, these improvements persist in out-of-distribution pathology datasets.
We hope that this improved generalization performance in imbalanced settings drives adoption of ML in real-world clinical applications.

\clearpage
{\small
\bibliographystyle{ieee_fullname}
\bibliography{egbib}
}

\end{document}